\documentclass[10pt,twocolumn,letterpaper]{article}

\usepackage[pagenumbers]{cvpr} 

\usepackage{graphicx}
\usepackage{amsmath}
\usepackage{amssymb}
\usepackage{booktabs}
\usepackage{balance}

\usepackage[pagebackref,breaklinks,colorlinks]{hyperref}

\usepackage[capitalize]{cleveref}
\crefname{section}{Sec.}{Secs.}
\Crefname{section}{Section}{Sections}
\Crefname{table}{Table}{Tables}
\crefname{table}{Tab.}{Tabs.}

\def\cvprPaperID{*****} 
\def\confName{CVPR}
\def\confYear{2023}

\begin{document}

\title{StillFast: An End-to-End Approach for \\ 
Short-Term Object Interaction Anticipation}

\author{Francesco Ragusa\\ \and Giovanni Maria Farinella \\ \and Antonino Furnari \\  \and
FPV@IPLab - University of Catania\\
Next Vision s.r.l. - Spin-off of the University of Catania\\
{\tt\small \{francesco.ragusa, giovanni.farinella, antonino.furnari\}@unict.it}
}
\maketitle

\begin{abstract}
Anticipation problem has been studied considering different aspects such as predicting humans' locations, predicting hands and objects trajectories, and forecasting actions and human-object interactions. In this paper, we studied the short-term object interaction anticipation problem from the egocentric point of view, proposing a new end-to-end architecture named StillFast. Our approach simultaneously processes a still image and a video detecting and localizing next-active objects, predicting the verb which describes the future interaction and determining when the interaction will start. Experiments on the large-scale egocentric dataset EGO4D \cite{Ego4D2022CVPR} show that our method outperformed state-of-the-art approaches on the considered task. Our method is ranked first in the public leaderboard of the EGO4D short term object interaction anticipation challenge 2022. Please see the project web page for code and additional details: \url{https://iplab.dmi.unict.it/stillfast/}.

\end{abstract}

\section{Introduction}
\label{sec:intro}

Anticipating human behavior in the near future from the first person (egocentric) point of view allows to build intelligent systems able to support users in different domains.
Anticipation is crucial in scenarios where one needs to react before actions are executed, such as in autonomous driving, where a vehicle has to anticipate pedestrians' trajectories before they even begin crossing the street \cite{Li_2022_CVPR, Mangalam2019DisentanglingHD}, in kitchens where smartglasses could alert the users when they are about to touch the hot stove \cite{Damen2018EPICKITCHENS,Damen2020RESCALING} or in the industrial domain, where an intelligent helmet can improve the worker's safety alerting them in case of a dangerous interaction \cite{ragusa2022meccano, VISAPP_2023_ENIGMA_CR}.

Previous works have investigated different forms of anticipation tasks, including next-active object predictions \cite{Furnari2017NextactiveobjectPF, JIANG2021212, ragusa2022meccano, Dessalene_forecasting_contact, Ego4D2022CVPR}, predicting future actions \cite{furnari2019rulstm, furnari2020rulstm, Gao2017REDRE,slowfast_rulstm_ballan, Vondrick2015AnticipatingVR, Felsen_what_will_happen_17, Roy_2022_WACV}, forecasting human-object interactions \cite{liu_forecasting_HOI}, predicting future hands \cite{Fan2017ForecastingHA, Jia2022GenerativeAN} or user trajectory prediction \cite{Park2016EgocentricFL}.

\begin{figure*}[t]
    \centering
    \includegraphics[width=\linewidth]{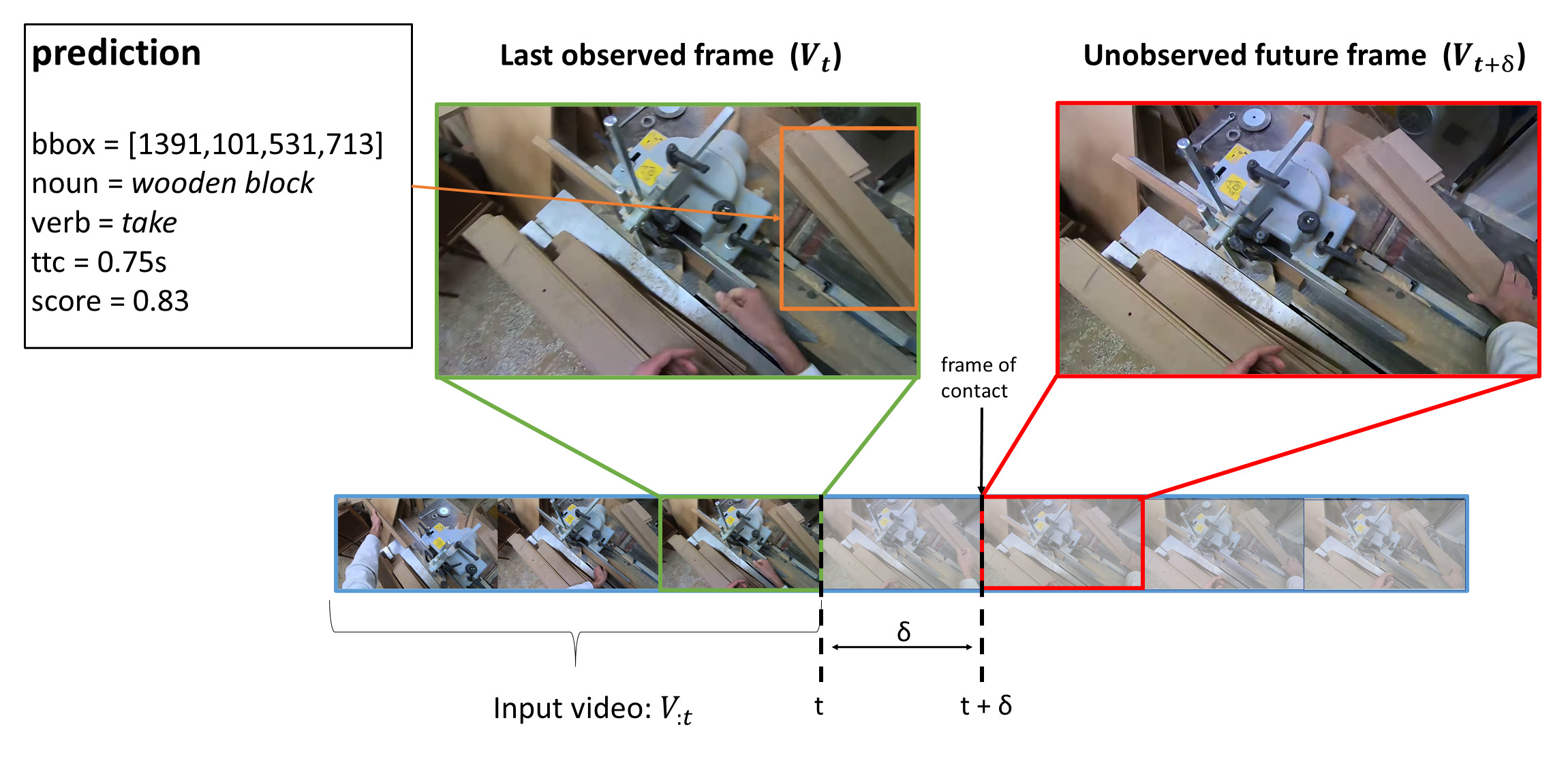}
    \caption{Short-term object interaction anticipation task. Models can process a video $V$ up to time $t$ (denoted as $V_{:t}$) predicting the bounding box and the class related to the next-active objects, the verb which describes the future interaction, a real number which indicating when the interaction will happen ($t+\delta$) and a score. $\delta$ represents the time interval between the last observable frame $V_t$ and the frame of contact at time $t+\delta$.}
    \label{fig:task}
\end{figure*}

In this paper we address the problem of short-term object interaction anticipation \cite{Ego4D2022CVPR} which consists in detecting and localizing the next-active objects in the scene, predicting the verb which describes the interaction and determining when the interaction will start. This task has been studied considering multiple domains \cite{Ego4D2022CVPR} as well as focusing specifically on the industrial scenario \cite{ragusa2022meccano}.
We define the task as proposed in \cite{Ego4D2022CVPR}. Given a video $V$ and a timestamp $t$, models can process the video up to time $t$ (denoted as $V_{:t}$) and are required to output a set of future object interaction predictions which will happen after a time interval $\delta$. Each prediction consists in:
\begin{itemize}
\item A bounding box localizing the future interacted object (also referred to as next-active object);
\item A noun label describing the class of the detected object (e.g., ``wooden block'');
\item A verb label describing the interaction which will take place in the future (e.g., ``take'');
\item A real number indicating the ``time to contact'', i.e., the time in seconds between the current timestamp and the beginning of the interaction (e.g., $0.75s$);
\item A confidence score used to rank future predictions for evaluation.
\end{itemize}
Figure~\ref{fig:task} illustrates the considered task.

Current state-of-the-art works \cite{Ego4D2022CVPR, ragusa2022meccano} have addressed this task in two steps: 1) an object detector detects and recognizes next-active objects in still frames, 2) a 3D network predicts the verb and the time to contact analyzing a video segment. Also in \cite{feichtenhofer2018slowfast}, the authors figured out the action detection task on AVA dataset considering a two steps approach.
Since past works showed that composite methods have been outperformed by end-to-end methods \cite{girshick2014rich, girshick2015fast, ren2015faster}, we believe that a similar behavior could be obtained even in the considered task.
Therefore, we propose \textit{StillFast} network. Similarly to SlowFast networks \cite{feichtenhofer2018slowfast}, StillFast has two branches which simultaneously process two versions of the input video. The “still” branch processes a high resolution still image, i.e., a video with a high spatial resolution, but low temporal resolution (a single frame), whereas the  “fast” branch processes a video with a low spatial resolution, but a high temporal resolution (different frames). Our method can be trained end-to-end in a single stage increasing the training speed over traditional two-branches approaches \cite{Ego4D2022CVPR}.

Experiments on the large-scale dataset of egocentric videos EGO4D \cite{Ego4D2022CVPR} show that the proposed method outperforms state-of-the-art approaches highlight that the proposed architecture benefits from the unified branches trained simultaneously.
Additionally, we performed an ablation study to assess the impact of each component of the proposed method on the overall performance. To encourage future research in the field, we release the code implementing the proposed approach at: \url{https://iplab.dmi.unict.it/stillfast/}.

The contributions of this work are as follows: 1) we proposed a new approach which is able to simultaneously processes a \textit{still image} and a \textit{fast video}, 2) by performing experiments with state-of-the art approaches and several ablation, we show the effectiveness of the proposed design, 3) we release the source code of the proposed approach as an extensible framework to encourage future research.
\section{Related Work}
\label{sec:relatedwork}
Our work is related to past research on anticipation considering both third and first-person point of view.

\begin{figure*}[t]
    \centering
    \includegraphics[width=\linewidth]{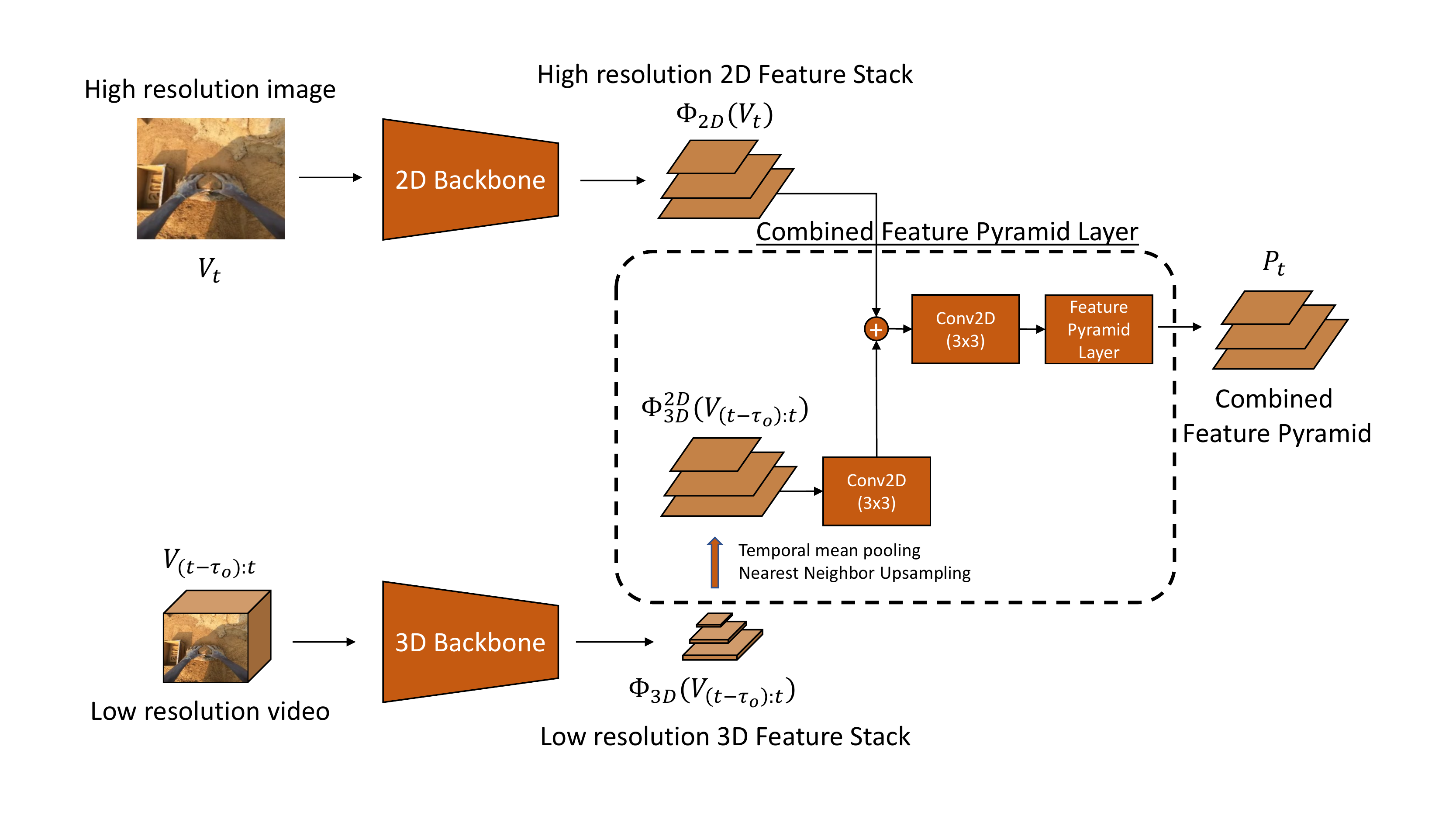}
    \caption{StillFast is composed of a two-branch backbone. Given an input video $V$ and a timestamp $t$, the proposed model takes as input a high resolution frame $V_t$ (top) and a low resolution video $(V_{(t-\tau_o):t})$ (bottom).  A 2D Backbone (``still'' branch) processes the high resolution frame $V_t$, producing a stack of 2D features $\Phi_{2D}(V_t)$. A 3D Backbone (``fast'' branch), processes a low resolution video $V_{(t-\tau_o):t}$ obtaining a stack of 3D features  $\Phi_{3D}(V_{(t-\tau_o):t})$. The Combined Feature Pyramid Layer is responsible to: 1) up-sample the stack of 3D features with nearest neighbor interpolation to match the spatial resolution of the 2D features and averages over the temporal dimension obtaining the $\Phi^{2D}_{3D}(V_{(t-\tau_o):t})$ features which have the same dimension of 2D features $\Phi_{2D}(V_t)$, 2) fuse these stack of features obtaining the final combined feature pyramid $P_t$. Before and after the sum operation we added 3x3 convolutional layers to remove artifacts introduced with the up-sampling and sum operations. }
    \label{fig:stillfast}
\end{figure*}

\subsection{Anticipation in Third Person Vision}
Different works aimed to predict future actions before they happen from a third person point of view\cite{Hierarchical_Repr_Savarese_14,Vondrick2015AnticipatingVR, Felsen_what_will_happen_17, Gao2017REDRE}.
The authors of \cite{Hierarchical_Repr_Savarese_14} proposed a new hierarchical representation called \textit{hierarchical movemes} which describes human movements considering multiple levels of granularity to infer future actions from still images or short video clips. The authors of \cite{Vondrick2015AnticipatingVR} explored the task of anticipating future actions and objects learning to predict future visual representations from unlabeled videos. The authors of \cite{Gao2017REDRE} proposed a reinforced encoder-decoder architecture (RED) for action anticipation which contains a reinforcement module to encourage the system to predict the correct action as early as possible. Also long-term action anticipation task has been explored in previous works \cite{Gong_2022_CVPR} with the aim to predict minutes-long sequences of future actions.
Even in the sport domain the problem of forecasting human moves has been explored \cite{Felsen_what_will_happen_17}. Indeed, the authors of \cite{Felsen_what_will_happen_17} proposed a generic framework to anticipate future events in team sports videos directly from visual inputs.
The anticipation problem has been explored even predicting the future location of users which allows to build an advanced surveillance systems able to predict people's activities\cite{Monti_2022_CVPR, Li_2022_CVPR}
or for autonomous vehicles to understand pedestrian intents to avoid accidents\cite{Mangalam2019DisentanglingHD,Mangalam_2021_ICCV}.
In particular, the authors of \cite{Mangalam2019DisentanglingHD} tackled the problem to jointly predicting the future spatial position and the body keypoints of pedestrians to have a deeper understanding of pedestrians behavior. The authors of \cite{Mangalam_2021_ICCV} proposed a factorized multimodal approach to predict long-term trajectories of the user focusing on RGB observations and past motion history.
Recently, the authors of \cite{Monti_2022_CVPR} studied the problem of using few input observations to predict accurate pedestrians position proposing a new teacher-student technique.

While prediction's tasks addressed from the third person point of view are useful in different scenarios, we focused on user-object interactions anticipation for which the egocentric point of view offers several advantages, therefore, we considered this anticipation problem relying on the large-scale egocentric dataset EGO4D\cite{Ego4D2022CVPR}.

\subsection{Anticipation in First Person Vision}
Different past works have explored problems related to anticipation from the first person point of view.
Furnari et al. \cite{furnari2019rulstm, furnari2020rulstm} proposed a model based on LSTM networks to encode past features and predict future actions from egocentric videos. The authors of \cite{slowfast_rulstm_ballan} extended RULSTM \cite{furnari2020rulstm} with a novel attention-based technique to consider simultaneously ``slow'' and ``fast'' features. The authors of \cite{multi_action_ant_zeyun23} proposed a novel transform-based fusion approach which combines multi-modal features (i.e., audio and visual) to predict future actions.
To better understand future human's behavior, past works also focused on predicting future hands and objects locations. The authors of \cite{liu_forecasting_HOI} presented a two-stream fully convolutional network to forecast the presence and location of hands and objects in egocentric videos.
Liu et al. \cite{liu_forecasting_HOI} focused on hands movements as visual representations to predict future interaction hotspots and future actions. 

While different tasks related to the anticipation problem have been deeply studied, the considered task have not been addressed systematically.

\subsection{Short-Term Object Interaction Anticipation}
The problem of Short-Term Object Interaction Anticipation has been addressed in different forms.
Some works studied the problem of predicting the objects which will be involved in a future human interaction (next-active objects). The authors of \cite{Furnari2017NextactiveobjectPF} have been the first to explore explicitly this task proposing to analyze objects trajectories. The authors of \cite{Dessalene_forecasting_contact} focused on the prediction of hand-objects contact representations to anticipate future actions. The authors of \cite{JIANG2021212} investigated the prediction of a visual attention probability map from images considering hands and objects features. The authors of \cite{ragusa2022meccano} presented a new egocentric dataset captured in a procedural scenario specifically annotated to address the next-active object detection task, which is tackled using simple object detectors.
All these works addressed different versions of the considered task, which made the proposed approaches difficult to compare and extend over. The first attempt to systematically study the problem of Short-Term Object Interaction Anticipation has been done by the authors of~\cite{Ego4D2022CVPR}, provided a formal definition of the task which includes the prediction of next-active objects and associated verbs and time to contact (see Figure~\ref{fig:task}).  
Despite this effort, few approaches have been proposed so far to tackle the task in the form defined in~\cite{Ego4D2022CVPR}. Among them, the baseline proposed in~\cite{Ego4D2022CVPR} consists of two components trained independently: A Faster R-CNN~\cite{ren2015faster} object detector to detect next-active objects in the last frame of the input video, and a SlowFast~\cite{feichtenhofer2019slowfast} action recognition model to attach each bounding box a verb and a time to contact prediction.
The model works as a two-branch network. The 2D Faster R-CNN model processes a high resolution image (the most recent frame of the input video) to predict next-active objects and their related classes. The high resolution is needed to have enough detail in order to detect objects at different scales. The 3D SlowFast model processes a low-resolution video clip ending at the current timestamp to predict verb labels and time-to-contact values. 
The approach is trained in two stages. First, the Faster RCNN model is trained using all training next-active object bounding box and class labels. This object detector is hence run over the training, validation and test data examples to complement them with a set of next-active object bounding box proposals with associated noun labels and detection scores. In a separate stage, a SlowFast model is trained to predict a verb label and a positive time-to-action real number for each bounding box proposal.
The authors of \cite{Chen2022InternVideoEgo4DAP} extended this approach by employing a
DINO\cite{Zhang2022DINODW} object detector to detect and recognize next-active objects in keyframes and a VideoMAE-pretrained transformer network~\cite{tong2022videomae} to extract features to predict verbs and  time to contact for each detected bounding box.

In this work, we propose an end-to-end approach specifically designed to anticipate the location of next-active objects, the verb which describes the future interaction and how soon the interaction will take place (time to contact). The proposed method can be trained in a single stage, reducing training times nd simplifying the research cycle.

\section{Still Fast Network}
\label{sec:method}

\begin{figure*}[t]
    \centering
    \includegraphics[width=\linewidth]{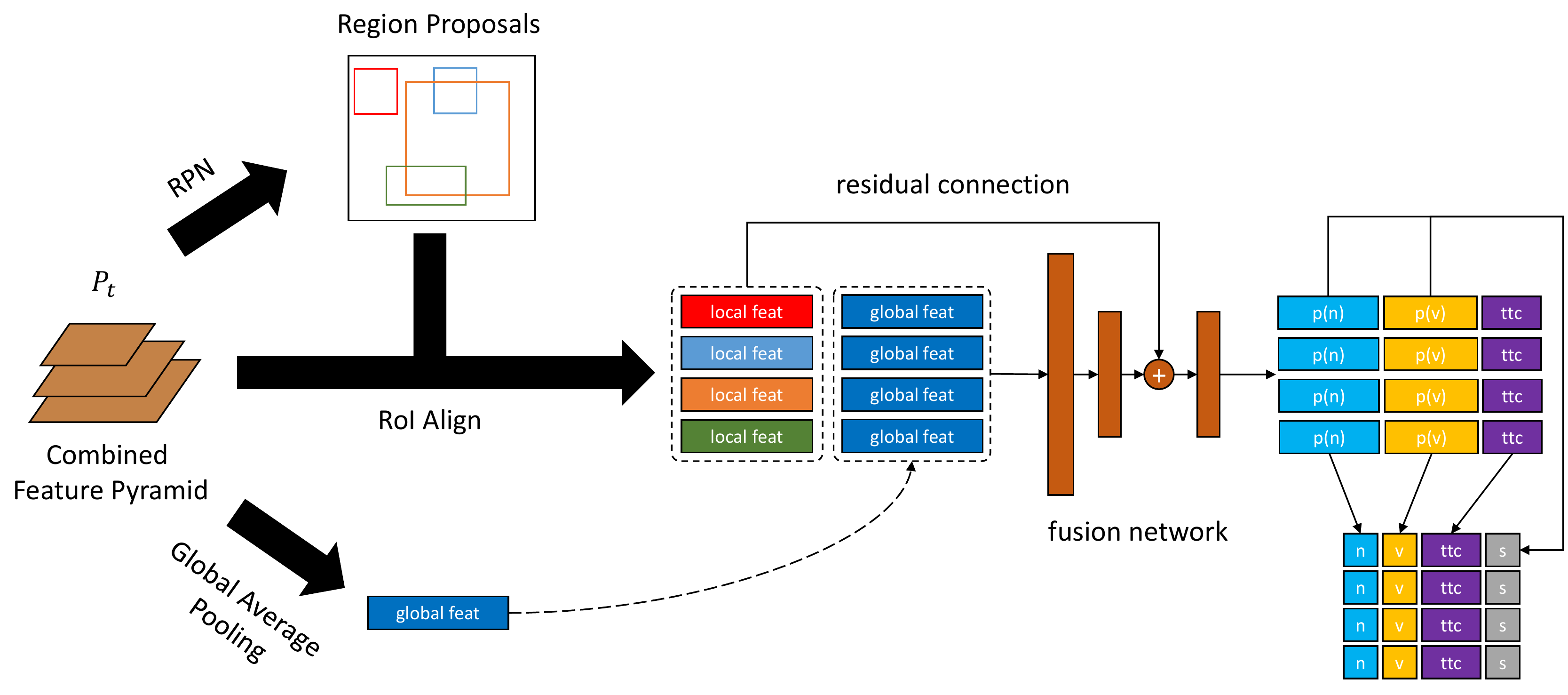}
    \caption{StillFast Prediction Head is based on the Faster R-CNN prediction head. From the Combined Feature Pyramid $P_t$ we obtain global and local features. Local features are obtained through a Region Proposal Network (RPN) which predicts region proposals, from which we compute local features through a RoI Align layer. Global features are obtained with a Global Average Pooling operation and are concatenated with local features. These features are fed in a fusion network and then are summed to the original local features through residual connections. These local-global representations are finally used to predict object (noun) and verb probability distributions and time-to-contact (ttc) through linear layers along with the related prediction score $s$.}
    \label{fig:head}
\end{figure*}

The proposed approach simultaneously processes a still image, i.e., a video with a high spatial resolution, but a low temporal resolution (a single frame) and a video with a low spatial resolution, but a high temporal resolution (different frames). We refer to the first branch as the ``still'' branch, as it processes a still image, whereas we refer to the second branch as the ``fast'' branch, as it processes a video with high temporal resolution. We hence term our model ``StillFast'' network. The model processes the two inputs simultaneously and can be trained end-to-end in a single stage, which increases training speed and allows to better optimize the feature extraction process.
The proposed method is comprised of two main components: a backbone with processes 2D (still image) and 3D (video) inputs and outputs a set of spatial features, and a prediction head, which takes as input the spatial features and outputs the detected next active objects together with the associated verbs and time to contact predictions.

The following subsections detail the main components of the proposed method.

\subsection{StillFast Backbone}
Figure~\ref{fig:stillfast} reports a diagram of the StillFast backbone. Given an input video $V$ and a timestamp $t$, the proposed model takes as input a high resolution frame $V_t$ sampled from video $V$ at time $t$ and a low resolution video $\mathcal{S}(V_{(t-\tau_o):t})$ which is obtained by spatially subsampling with function $\mathcal{S}$ the video $v_{(t-\tau_o):t}$ of length $\tau_o$ (observation time) starting at time $t-\tau_o$ and ending at time $t$. 

To process the input image and video simultaneously, the proposed model comprises a two-branch backbone. 

A 2D CNN $\Phi_{2D}$ (the ``still'' branch) processes the high resolution frame $V_t$ and produces a stack of 2D features at different spatial resolutions $\Phi_{2D}(V_t)$. The stack of features is obtained by collecting activations at the inner layers of the 2D network. As detailed later, we follow \cite{lin2017feature} to obtain the stack of features to produce a feature pyramid in order to enable multi-scale object detection as done in the standard~\cite{ren2015faster} architecture. 

A 3D CNN $\Phi_{3D}$ (the ``fast'' branch) processes the video $V_{(t-\tau_o):t}$ and outputs a stack of 3D features $\Phi_{3D}(V_{(t-\tau_o):t})$. We perform mid-level feature fusion by combining the 2D and the 3D feature stacks with a Combined Feature Pyramid Layer illustrated in the center part of Figure~\ref{fig:stillfast}. This layer first up-samples the 3D feature map with nearest neighbor interpolation to match the spatial resolution of the 2D features and averages over the temporal dimension to obtain the $\Phi^{2D}_{3D}(V_{(t-\tau_o):t})$ features, which now have the same shape as the 2D features $\Phi_{2D}(V_t)$. These features are hence passed through a 3x3 convolutional layer, summed to the 2D features $\Phi_{2D}(V_t)$ and then passed through another 3x3 convolutional layer. The rationale behind these 2D convolutional layers is to cope with artifacts introduced with the up-sampling and sum operations. The resulting feature map is fed to a standard Feature Pyramid Layer~\cite{lin2017feature} to obtain the final feature pyramid $P_t$.

In our experiment we use a ResNet-50 as 2D CNN and an X3D-M as 3D CNN. See the supplementary material for more details.

\subsection{Prediction Head}
\label{sec:head}
Figure~\ref{fig:head} shows a diagram of the proposed prediction head, which is based on the Faster R-CNN prediction head~\cite{ren2015faster} as implemented in Detectron2~\cite{wu2019detectron2}. A Region Proposal Network (RPN) predicts region proposals from the feature pyramid $P_t$. A RoiAlign layer is used to extract local features from the region proposals. We found it useful to enrich these local features with global representations of the scene. To do so, we apply global average pooling to the upper layer of the combined feature pyramid and obtain a global image representation, which we concatenate to each local feature extracted from the region proposals. The concatenated features are passed through a fusion network comprising two fully connected layers. The resulting representations are summed to the original local features through a residual connection. This allows to use global features to modulate the content of local features rather than to replace them. These fused local-global representations are used to predict object (noun) probability distributions and class-specific bounding box regression offsets using linear layers as in \cite{ren2015faster}. The predicted noun probability distributions $p(n)$ include a background class to reduce the magnitude of positive prediction probabilities in the case of uncertain predictions or proposals falling in background areas. The same features are used to predict a verb probability distribution $p(v)$ and time-to-contact TTC using linear layers. We include a background class in the verb prediction layer as well to further discard false positives when the verb cannot be reliably predicted. A softplus activation is used to predict positive TTC values. For each object proposal, we multiply the noun probability $p(n)$ by the predicted probability of the Top-1 verb, excluding the background class. This is done to make sure that the uncertainty in verb prediction influences the final prediction score $s=p(n)\cdot max_v\{p(v)\}$. The final predictions are obtained by considering objects with a prediction score larger than or equal to $0.05$. The model is trained end-to-end adding a cross entropy verb loss $\mathcal{L}_v$ and a smooth L1~\cite{ren2015faster} time-to-contact loss $\mathcal{L}_{ttc}$ to the standard Faster R-CNN losses. We weigh the $\mathcal{L}_v$ with $0.1$ and the $\mathcal{L}_{ttc}$ with $0.5$ in our experiments. See the supplementary material for more details.

\section{Experimental Settings}
\label{sec:experimental_sett}
In this section, we report details on the considered dataset and evaluation measures (Section~\ref{sec:dataset}) and the compared methods (Section~\ref{sec:methods}). Please see the supplementary material for the implementation details. 

\subsection{Dataset and Evaluation Measures}
\label{sec:dataset}
We performed experiments on the large-scale egocentric dataset EGO4D\cite{Ego4D2022CVPR}. 
We consider both the initial version of the dataset described in~\cite{Ego4D2022CVPR} (denoted as ``v1'' in this paper), and the recently released update of the dataset (denoted as ``v2''), described at this page: \url{https://ego4d-data.org/docs/updates/#v20-update}, following the official training/evaliuation/testing splits.
We focus on the subset of the EGO4D dataset which has been explicitly labeled for the Short-Term Object Interaction Anticipation task. Version v1 of this dataset consists in $120$ hours of annotated clips, including $27,801$ training, $17,217$ validation, and $19,780$ test examples, annotated with a taxonomy of $87$ noun and $74$ verb classes. Version v2 consists in $243$ hours of annotated clips, including $98,276$ training, $47,395$ validation, and $19,780$ test examples, annotated with a taxonomy of $128$ noun and $81$ verb classes.
It should be noted that v1 and v2 have different training and validation sets, but they share the same test set, which makes test results obtained on the two dataset comparable. Annotations for this shared test set are not publicly available and results can be obtained by sending predictions to an evaluation server\footnote{\url{https://eval.ai/web/challenges/challenge-page/1623/}}.

The evaluation has been performed using Top-K mean Average Precision, which does not penalize methods predicting up to K - 1 next-active objects which are not annotated as defined in \cite{Ego4D2022CVPR}. In particular, we evaluated methods with different Top-5 mAP measures, i.e., Top-5 mAP Noun, Top-5 mAP Noun+Verb, Top-5 mAP Noun+TTC and Top-5 mAP Noun+Verb+TTC which we named Top-5 mAP Overall, to assess the ability of the model to anticipate next-active object interactions considering different levels of granularity (i.e., nouns, verbs, time-to-contact). We used $K=5$ as defined in \cite{Ego4D2022CVPR}.

\subsection{Compared Methods}
\label{sec:methods}
We compare our method with respect to different approaches which addressed the considered Short-Term Object Interaction Anticipation problem.
\begin{itemize}
    \item Faster R-CNN + Random\cite{Ego4D2022CVPR}: uses Faster R-CNN to detect and recognize next active objects, then predicts verbs and time to contact randomly following the distributions of training labels;
    \item Faster R-CNN + SlowFast\cite{Ego4D2022CVPR}: the baseline defined in \cite{Ego4D2022CVPR} composed of two components trained independently. A Faster R-CNN \cite{ren2015faster} object detector which detects and recognizes next-active objects and a SlowFast \cite{feichtenhofer2018slowfast} 3D network to predict for each bounding box a verb and a time to contact;
    \item InternVideo\cite{Chen2022InternVideoEgo4DAP}: follows the same two-stage recipe of Faster R-CNN + SlowFast. A DINO\cite{Zhang2022DINODW} object detector is used to predict next-active objects and a VideoMAE-pretrained trasnformer video backbone~\cite{tong2022videomae} to predict verb and time to contact for each detected next-active object.
    \item Faster R-CNN + Features: a Faster R-CNN object detector is used to detect and recognize next active objects, then pre-extracted Ominvore features \cite{Girdhar2022OmnivoreAS} are used to predict the associated verb and time to contact. This baseline has been proposed as a quickstart approach with the EGO4D dataset. In our experiments, we consider the results reported in this document: \url{https://colab.research.google.com/drive/1Ok_6F1O6K8kX1S4sEnU62HoOBw_CPngR}.

\end{itemize}

\section{Results}
\label{sec:results}
This section compares the proposed StillFast method to competitors (Section~\ref{sec:sota}) and analyses the contribution of the different components of the approach to performance (Section~\ref{sec:ablation}).

\begin{table}[t]
\centering
\resizebox{\columnwidth}{!}{%
    \begin{tabular}{clc|cccc}
    \hline
        \textbf{Set} & \textbf{Method} &\textbf{Ver} & \textbf{Noun} & \textbf{N+V} & \textbf{N+TTC} & \textbf{Overall} \\
        \hline
        Val & FRCNN+Rnd.~\cite{Ego4D2022CVPR} & v1 & 17.55 & 1.56 & 3.21 & 0.34\\
        Val & FRCNN+SF~\cite{Ego4D2022CVPR} & v1 &\textbf{17.55} & 5.19 & \textbf{5.37} & 2.07 \\
        Val & StillFast (ours) & v1 &16.20 & \textbf{7.47} & 4.94 & \textbf{2.48} \\
        \hline
        Val & FRCNN+SF~\cite{Ego4D2022CVPR} & v2 & \textbf{21.00} & 7.45 & 7.04 & 2.98 \\
        Val & StillFast (ours) & v2 & 20.26 & \textbf{10.37} & \textbf{7.16} & \textbf{3.96} \\
        \hline
        Test & FRCNN+Rnd.~\cite{Ego4D2022CVPR} &v1 & 20.45 & 2.22 & 3.86 & 0.44 \\
        Test & FRCNN+SF~\cite{Ego4D2022CVPR} &v1 & 20.45 & 6.78 & 6.17 & 2.45 \\
        Test & FRCNN+Feat. &v1 & 20.45 & 4.81 & 4.40 & 1.31\\
        Test & InternVideo~\cite{Chen2022InternVideoEgo4DAP} &v1 & \textbf{24.60} & 9.18 & \textbf{7.64} & 3.40\\
        Test & StillFast (ours) & v1 & 19.51 & \textbf{9.95} & 6.45 & \textbf{3.49} \\
        \hline
        Test & FRCNN+SF~\cite{Ego4D2022CVPR} & v2 & \textbf{26.15} & 9.45 & 8.69 & 3.61 \\
        Test & StillFast (ours) & v2 & 25.06 & \textbf{13.29} & \textbf{9.14} & \textbf{5.12} \\
        \hline
    \end{tabular}
    }
    \caption{Results\% in Top-5 mean Average Precision on the validation and test sets of EGO4D v1 and v2. In the header of the table, Ver stands for Version, V+N stands for Verb + Noun and N+TTC stands for Noun + Time to Contact. Best results per column within a section of comparable results (horizontal lines) are reported in bold.}
    \label{tab:results}
\end{table}

\subsection{Comparison with the State of the Art}
\label{sec:sota}
Table~\ref{tab:results} reports the results of the compared methods the validation and test sets of both v1 and v2 of the EGO4D dataset~\cite{Ego4D2022CVPR} using the aforementioned Top-5 mAP evaluation measure.
As can be noted from Table~\ref{tab:results}, the proposed method improves over the FRCNN+Rnd. and FRCNN+SF baselines by significant margins in the v1 validation set on verb-related metrics: $7.47\%$ vs $5.19\%$ Noun+Verb Top-5 mAP ($+2.28\%$), and $2.48\%$ vs $2.07\%$ Overall Top-5 mAP ($+0.41\%$). 
These results suggest that the ability of the proposed approach to process images and video with a unified backbone and accounting for the uncertainty in verb prediction as described in Section~\ref{sec:head} is beneficial to performance. Indeed, it should be noted that both are based on components with similar performance: a ResNet-50 2D backbone in both cases, a SlowFast 3D backbone for FRCNN+SF and an X3D-M backbone for StillFast. 
On the downside, the proposed StillFast approach achieves worse performance on Noun Top-5 mAP ($16.20\%$ vs $17.55\%$, hence $-1.35\%$) and on Noun + TTC Top-5 mAP ($4.94\%$ vs $5.37\%$, hence $-0.43\%$). We speculate that this is due to training instabilities due to the multi-task nature of our training procedure, as compared with FRCNN+SF, which is trained in two separate stages (Section~\ref{sec:ablation} further analyses performance when multi-tasking is reduced within our architecture). This seems to be mitigated when more training data is available. Indeed, when training on v2 (fourth and fifth row of Table~\ref{tab:results}), StillFast consistently outperforms FRCNN+SF on Noun + Verb, Noun + TTC and Overall Top-5 mAP, while the loss of performance on Noun Top-5 mAP is much smaller ($20.26\%$ vs $21.00\%$, hence only $-0.74\%$).

The advantages of StillFast are more evident in the test set, which suggests better generalization of the proposed approach. In v1, the proposed method outperforms FRCNN+SF by $+3.17\%$ ($9.95\%$ vs $6.78\%$) according to Noun+Verb Top-5 mAP, $+0.28\%$ ($6.45\%$ vs $6.17\%$) according to Noun+TTC Top-5 mAP, and $+1.04\%$ ($3.49\%$ vs $2.45\%$) according to Overall Top-5 mAP. 
Noun Top-5 mAP results are still lower, but closer to the ones of FRCNN+SF ($19.51\%$ vs $20.45\%$, hence $-0.95\%$). Also in the test set, using more training data brings larger performance margins to StillFast. In v2 test set, the proposed approach surpasses FRCNN+SF by $+3.84\%$ ($13.29\%$ vs $9.45\%$) according to Noun + Verb Top-5 mAP, $+0.45\%$ ($9.14\%$ vs $8.69\%$) according to Noun + TTC Top-5 mAP, and $+1.51\%$ ($5.12\%$ vs $3.61\%$) according to Overall Top-5 mAP. If we consider that v1 and v2 share the same test set, StillFast jumps from a $9.95\%$ of v1 to $13.29\%$ of v2 Noun + Verb Top-5 mAP and from $3.49\%$ of v1 to $5.12\%$ of v2 Overall mAP, by merely adding more training data, which suggests that the proposed approach can scale in the presence of larger datasets.

In the v1 test set, we are able to also compare StillFast with other methods for which results are publicly available, including InternVideo~\cite{Chen2022InternVideoEgo4DAP} and FRCNN+Feat. StillFast outperforms FRCNN+Feat. on all evaluation measures except Noun Top-5 mAP, where it achieves slightly worse performance ($19.51\%$ vs $20.45\%$). Note that FRCNN+Feat. uses the same Faster RCNN object detector component as FRCNN+SF and is trained in two stages, hence similar considerations as the ones made while comparing the proposed approach with FRCNN+SF apply here as well.
StillFast still outperforms InternVideo with respect to Noun+Verb Top-5 mAP ($9.95\%$ vs $9.18\%$) and Overall Top-5 mAP ($3.49\%$ vs $3.40\%$) despite being based on less advanced components (StillFast relies on a Faster R-CNN object detector, while InternVideo relies on DINO~\cite{Zhang2022DINODW}, StillFast relies on an X3D video backbone, while InternVideo relies on Video-MAE~\cite{tong2022videomae} pre-trained transform-based video backbones) and being trained end-to-end (InternVideo is trained in two stages as FRCNN+SF). We leave the integration of higher performing components in StillFast to future works.

\begin{figure*}[t]
    \centering
    \includegraphics[width=0.24\linewidth]{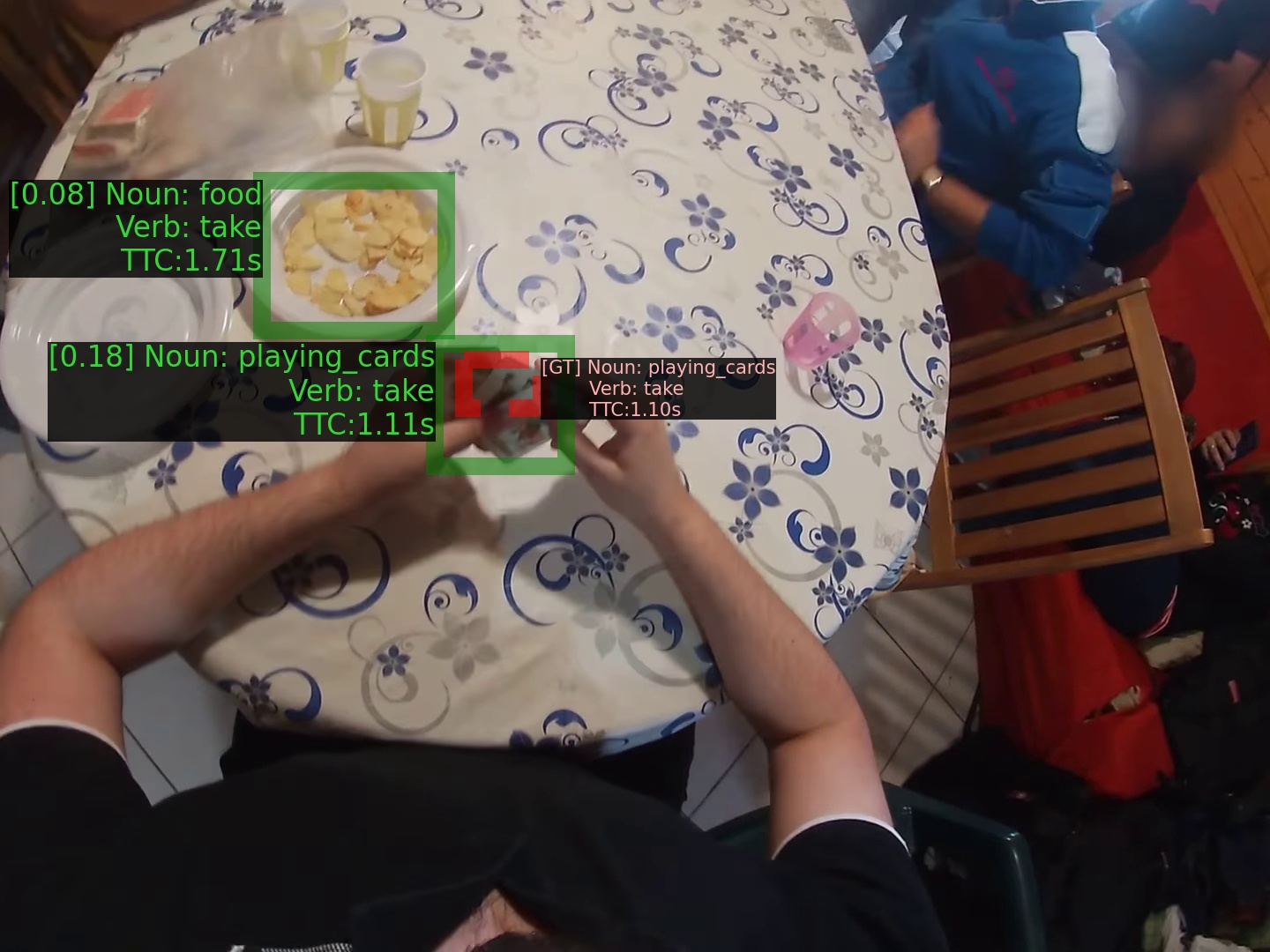} \hfill
    \includegraphics[width=0.24\linewidth]{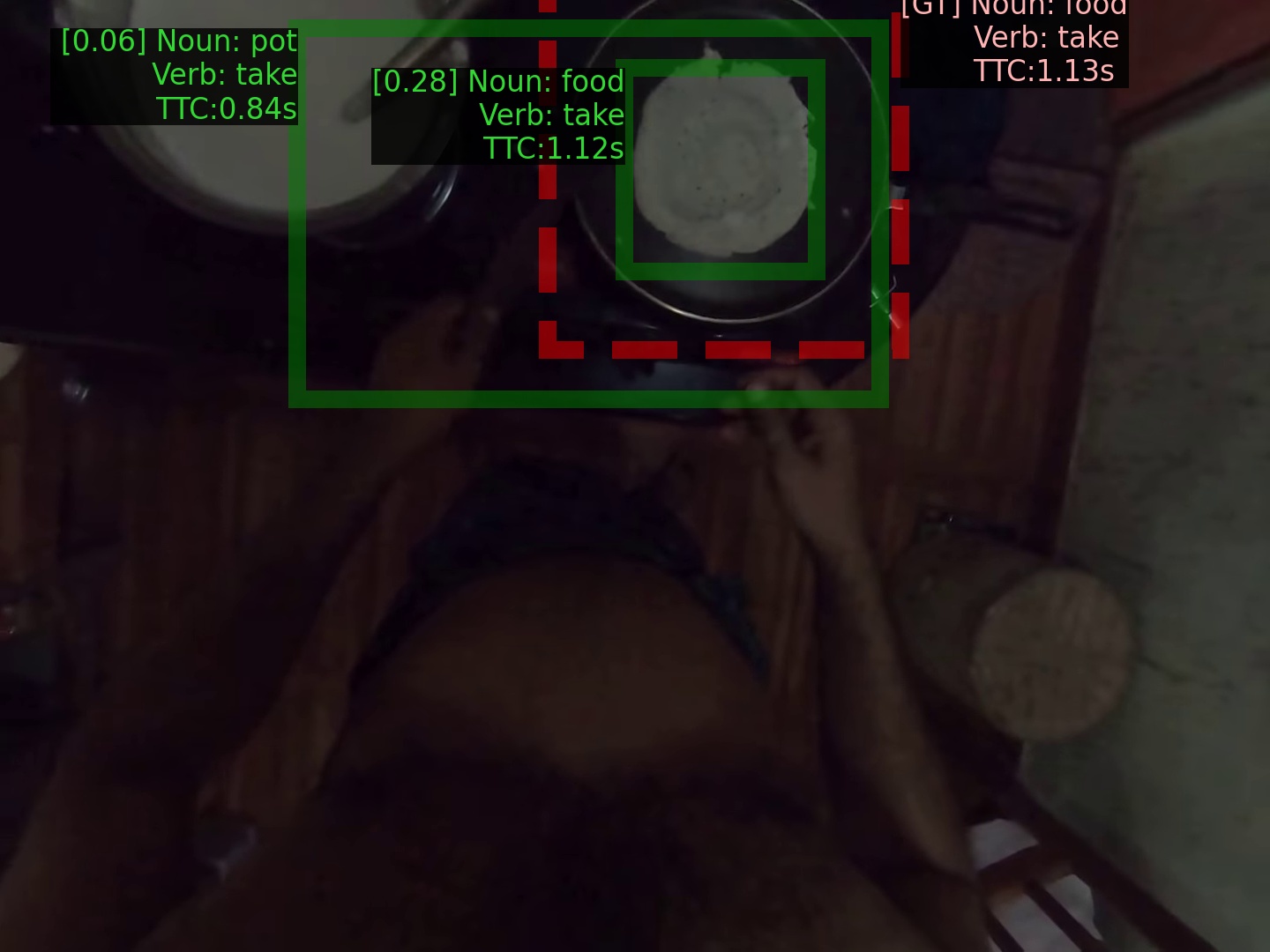} \hfill
    \includegraphics[width=0.24\linewidth]{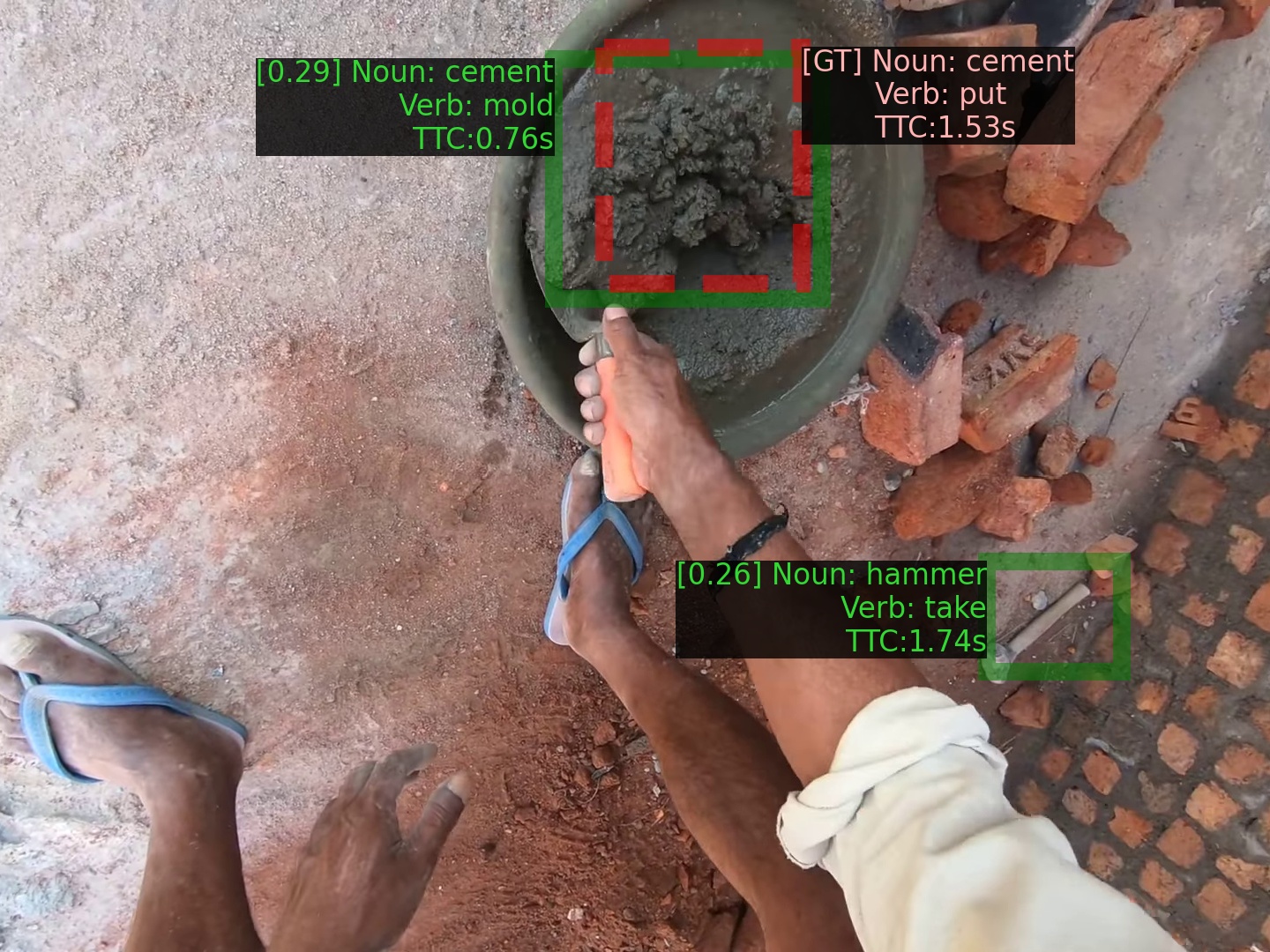} \hfill
    \includegraphics[width=0.24\linewidth]{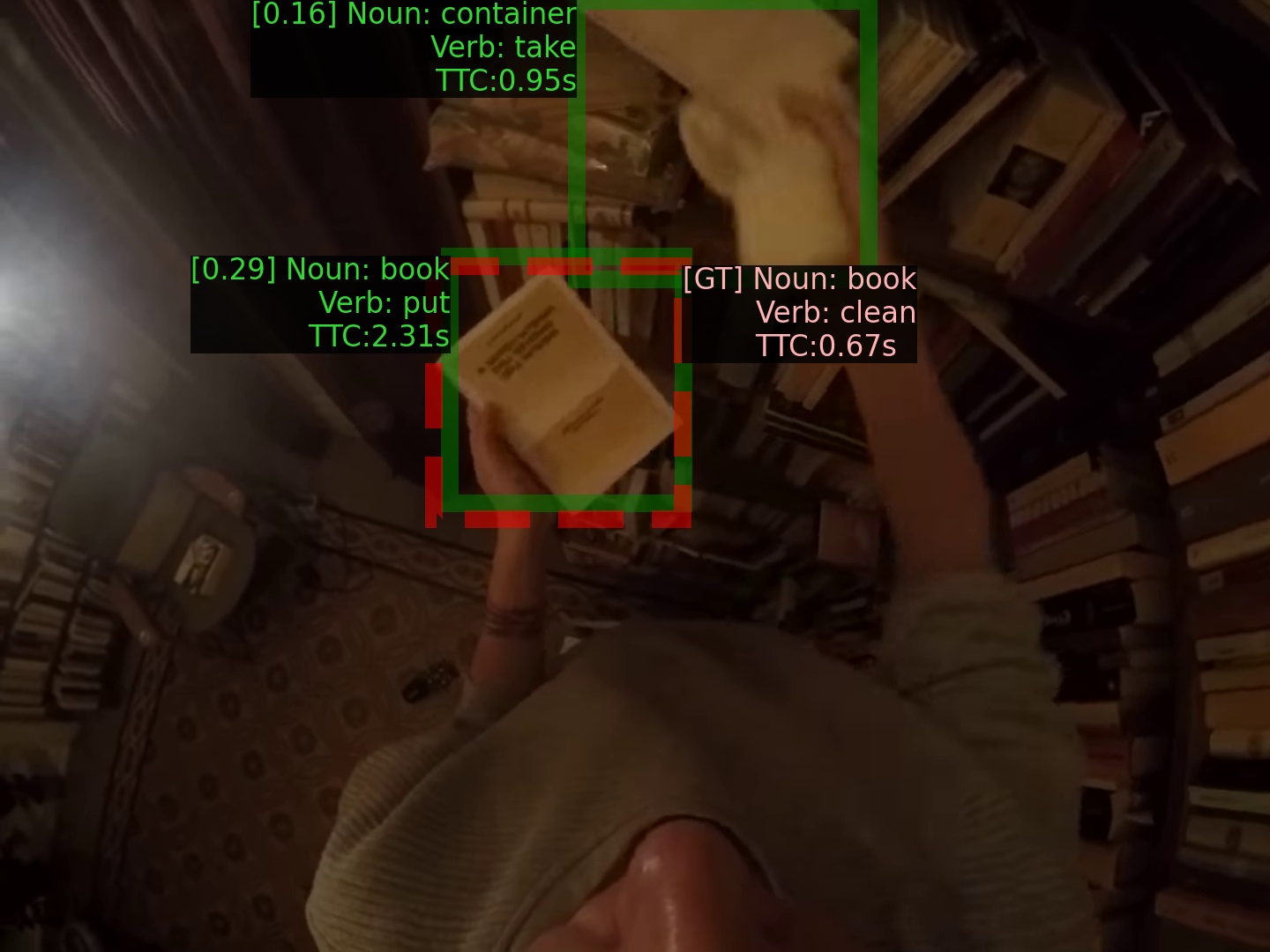}
    \caption{Two success examples (left) and two failure cases (right).}
    \label{fig:examples}
\end{figure*}

Figure~\ref{fig:examples} reports two success (left) and two failure examples (right). The model struggles with uncertain future actions (``put cement'' vs ``mold cement'') and unusual actions (``clean book'' vs ``put book'').

\subsection{Ablation study}
\label{sec:ablation}
In this section, we further investigate other aspects of the proposed StillFast architecture including the verb/noun prediction trade-off introduced by the prediction head described in Section~\ref{sec:head}, and the contribution of different architectural choices to the overall performance. The tables in this section report validation results of experiments performed on EGO4D v1 data. All results are in Top-5 mAP\%.

\begin{table}[t]
    \centering
    \resizebox{\columnwidth}{!}{%
    \begin{tabular}{l|ccccc}
    \hline
        \textbf{Method} & \textbf{Noun} & \textbf{N+V} & \textbf{N+TTC} & \textbf{Overall} \\
        \hline
        FRCNN+SF~\cite{Ego4D2022CVPR} & 17.55 & 5.19 & \textbf{5.37} & 2.07 \\
        \hline
        Nouns Only & \textbf{19.69} & - & - & - \\
        Standard Head & 18.42 & 6.39 & 5.28 & 2.17 \\
        Proposed Head & 16.20 & \textbf{7.47} & 4.94 & \textbf{2.48} \\
        \hline
    \end{tabular}
    }
    \caption{Comparison of the performance of different heads.}
    \label{tab:ablation}
\end{table}

\begin{table}[t]
    \centering
    \resizebox{\columnwidth}{!}{%
    \begin{tabular}{l|ccccc}
    \hline
        \textbf{Method} & \textbf{Noun} & \textbf{N+V} & \textbf{N+TTC} & \textbf{Overall} \\
        \hline
        Proposed Head & 16.20 & \textbf{7.47} & \textbf{4.94} & 2.48 \\
        \hline
        -global features& \textbf{16.90} & 5.77 & 4.60 & 1.94 \\
        -res. connections & 15.44 & 6.40 & 4.80 & 2.25 \\
        -verb-noun product & 14.95 & 6.29 & 4.28 & 1.78 \\
        \hline
        Sum Fusion                  & 15.26                             & 6.68                                   & 4.93                                  & \textbf{2.49}                                 \\ 
        \hline
    \end{tabular}
    }
    \caption{Ablation study of the proposed head.}
    \label{tab:ablation_head}
\end{table}

\textbf{Prediction Head} Table~\ref{tab:ablation} compares the results obtained by the proposed model when different prediction heads are used. The results of FRCNN+SF are also reported for reference. ``Nouns Only'' predicts only next-active objects and does not predict any verb or time to contact, ``Standard Head'' refers to the standard prediction head proposed in~\cite{Ego4D2022CVPR} which uses a SlowFast model to attach a verb and a time to contact prediction to each detected bounding box. ``Proposed Head'' refers to the head described in~\ref{sec:head}, including global features and accounting for verb prediction uncertainty.
As can be noted, predicting only nouns outperforms the FRCNN+SF baseline by $+2.05\%$ ($19.69\%$ vs $17.55\%$) with respect to Noun Top-5 mAP. Adding verb and time to contact prediction with a standard head decreases Noun Top-5 mAP performance to $18.42\%$, which still outperforms FRCNN+SF ($17.55\%$), while obtaining better or comparable performance on the other metrics. This suggests that the proposed approach is penalized by the training instabilities caused by multi-tasking (i.e., predicting simultaneously nouns, verbs and time to contact). 
The proposed head achieves the best results in terms of Noun+Verb Top-5 mAP and Overall performance thanks to the use of global features and verb-noun score product, which effectively accounts for the uncertainty in the prediction of verbs. 
Table~\ref{tab:ablation_head} assesses the impact of the main components of the proposed head, comparing the performance of the overall module with versions in which we remove global features, residual connections, and the verb-noun score product module. We also compare the head with a version which replaces the proposed global-local fusion mechanism by a simple sum (last row). As can be observed, the proposed architecture obtains overall the best results. Sum fusion achieves slightly better overall Top-5 mAP ($+0.01\%$), but lower results according to the other metrics, for which we prefer the proposed concatenation + residual connection design.

\begin{table}[t]
\centering
\resizebox{\columnwidth}{!}{%
\begin{tabular}{lcccc}
\multicolumn{1}{l|}{}                            & \multicolumn{1}{l}{\textbf{Noun}} & \multicolumn{1}{l}{\textbf{N+V}} & \multicolumn{1}{l}{\textbf{N+TTC}} & \multicolumn{1}{l}{\textbf{Overall}} \\ \hline
\multicolumn{1}{l|}{Proposed backbone}                                       & \textbf{16.20}                    & \textbf{7.47}                          & \textbf{4.94}                         & \textbf{2.48}               \\        
\hline
\multicolumn{1}{l|}{w/o 3D backbone}             & 14.54                             & 6.17                                   & 4.10                                  & 1.82                                 \\
\multicolumn{1}{l|}{w/o conv. block post sum}    & 15.13                             & 6.79                                   & 4.80                                  & 2.25                                 \\
\hline
\multicolumn{1}{l|}{post-pyramid fusion}    & 15.01 & 6.74 & 4.69 & 2.34                      \\

\hline
\end{tabular}
}
 \caption{Ablation study considering the contribution of each component of the proposed architecture.}
    \label{tab:ablation_component}
\end{table}

\textbf{Backbone} Table~\ref{tab:ablation_component} shows the impact of the different components in the backbone when the proposed head is used. 
We observe that without a 3D branch (second row) the performance of the method drops according to all mAP measures by significant margins (e.g., $1.82\%$ vs $2.48\%$ Overall Top-5 mAP). This suggests that the proposed backbone design succesfully combines 3D and 2D features to anticipate future interactions. The third row shows that removing 2D convolutional blocks after the sum operation in the combined feature pyramid layer negatively affect performance (e.g., $2.25\%$ vs $2.48\%$ Top-5 Overall mAP). We speculate that the use of these layers allows to deal with artifacts potentially introduced in the upsampling and sum operations. The last row reports the performance of an alternative design of the network which attaches a 2D feature pyramid to the 2D network and a 3D feature pyramid to the 3D network and then fuses the resulting feature pyramids. As can be noted, this design is less effective than the proposed one, with a Top-5 Overall mAP score equal to $2.34\%$ (vs $2.48\%$ of the proposed design).

\section{Conclusion}
\label{sec:conclusion}
In this paper, we have presented StillFast, an end-to-end approach to tackle the Short Term Object Interaction Anticipation task. Differently from previous methods, StillFast is designed to process video and image inputs simultaneously and encourage future research in this field by providing a practical framework designed to be easily extensible. To facilitate future research, we will release the source code which can be used to replicate the experiments reported in this paper. While the proposed method achieves promising performance EGO4D, we are aware of some of the limitations of the approach, which we leave to future works. Specifically, the current implementation of the method is based on convolutional 2D and 3D backbones, while benefits may arise from the use of more recent transformer-based components as done in previous works~\cite{Chen2022InternVideoEgo4DAP}. Moreover, the current end-to-end training procedure seems to be penalized by the multi-tasking arising from predicting noun, verb and time to contact within a single model. We trust that future works will be able to address such limitations and extend on the proposed work. Future investigations may also explore the potential benefit of the proposed architecture on other tasks requiring the simultaneous analysis of images and videos, such as for instance action detection.

\section*{\uppercase{Acknowledgements}}
This research is supported by Next Vision\footnote{Next Vision: https://www.nextvisionlab.it/} s.r.l., by the project Future Artificial Intelligence Research (FAIR) – PNRR MUR Cod. PE0000013 - CUP: E63C22001940006 and by the project MISE - PON I\&C 2014-2020 - Progetto ENIGMA - Prog n. F/190050/02/X44 – CUP: B61B19000520008.

\section*{\uppercase{Supplementary Material}}
\label{sec:supp_material}
This document is intended for the convenience of the reader and reports additional information about the 
implementation details. This supplementary material is related to the following submission:\\

F. Ragusa, G. M. Farinella, A. Furnari. StillFast: An End-to-End Approach for Short-Term Object Interaction Anticipation. Proceedings of the IEEE/CVF Conference on Computer Vision and Pattern Recognition (CVPR) Workshops. 2023. 

\section{Implementation Details}
\label{sec:implementation}
 
We train the model on the training data, whereas we use the validation set to choose the best performing checkpoint according to the overall mAP measure. We then report the results of such model on both the validation and the test set. At training time, we follow the standard Faster R-CNN multi-scale procedure and feed high-resolution images with a short side in the range $[640, 672, 704, 736, 768, 800]$ and a maximum long side of $1333$. As a result, we obtain a still image with height $H$. The low resolution video is obtained by re-scaling the input high resolution video with linear interpolation in such a way that the height of re-scaled video is equal to $h=\alpha \times H$. We set $\alpha=0.32$ in all our experiments. In this way, a still image of height $H=800$ pixels will correspond to a video of height $h=256$ pixels, which is a standard resolution for video backbones. At test time, we feed to the networks still images of height $H=800$ pixels and videos of height $h=256$ pixels. We sample video clips of $16$ frames with a sampling stride of $1$ frame. During training, we weigh the $\mathcal{L}_v$ loss with $0.1$ and the $\mathcal{L}_{ttc}$ loss with $0.5$. The 2D backbone is a ResNet-50 architecture. The weights of this backbone and the ones of the standard feature pyramid layer are initialized from a Faster R-CNN model pre-trained on the COCO dataset~\cite{lin2014microsoft}. The 3D network is an X3D-M model~\cite{feichtenhofer2020x3d} pre-trained on Kinetics~\cite{carreira2017quo}. The global-local fusion network included in the prediction head has two connected layers with a ReLU activation in between. The first layer maps features from $256+1024$ features to $1024$ features, whereas the second layer maps features from $1024$ to $1024$ dimensions. The weights of the local-global module are initialized randomly. The model is trained with a base learning rate of $0.001$ and a weight decay of $0.0001$. The learning rate is lowered by a factor of $10$ after $15$ and $30$ epochs. The model is trained in half precision on four NVIDIA V100 GPUs with a batch size of $14$. The convolutional layers included in the Combined Feature Pyramid Layers are randomly initialized and have a $3\times3$ kernel with a padding equal to $1$. The first convolutional layer (pre-sum) maps features from the numbers of channels of the 3D network ($[24, 48, 96, 192]$) to numbers of channels of the 2D network ($[256, 512, 1024, 2048]$), whereas the second convolutional layer (post-sum) maps features from the number of channels of the 2D network ($[256, 512, 1024, 2048]$) to the same number of channels. The standard feature pyramid layer maps features to $256$ channels. Following the 2D and 3D backbone branch initialization, input still images are normalized with $[0.485, 0.456, 0.406]$ means and $[0.229, 0.224, 0.225]$ standard deviations, whereas the input videos are normalized with $[0.45, 0.45, 0.45]$ means and $[0.225, 0.225, 0.225]$ standard deviations.

{\small
\balance
\bibliographystyle{ieee_fullname}
\bibliography{egbib}
}

\end{document}